\documentclass[preprint,12pt]{elsarticle}

\usepackage[utf8]{inputenc}
\usepackage[dvipsnames]{xcolor}
\usepackage{tikz}
\usepackage{url}
\usepackage{amsmath}
\usepackage[ruled,vlined]{algorithm2e}
\usepackage{mathtools}

\definecolor{royalblue}{cmyk}{1,0.50,0,0} 
\definecolor{cerulean}{cmyk}{0.94,0.11,0,0} 
\definecolor{violet}{cmyk}{0.79,0.88,0,0}
\definecolor{royalred}{rgb}{0.6,0.11,0.19}

\newcommand{\approach}{TiWSiF \ }
\newcommand{\tree}[1]{{isolation tree}}
\newcommand{\forest}[1]{{isolation forest}}
\newcommand{\ap}{\bar{p}}
\usepackage{subfig}
\usepackage{booktabs}

\usepackage{amssymb}

\journal{Information Sciences}

\begin{document}

\begin{frontmatter}



\title{TiWS-iForest: Isolation Forest in Weakly Supervised and Tiny ML scenarios}


\author[inst1]{Tommaso Barbariol}

\affiliation[inst1]{organization={Department of Information Engineering, University of Padova},
            addressline={Via Giovanni Gradenigo 6}, 
            city={Padova},
            postcode={35131}, 
            state={PD},
            country={Italy}}

\author[inst1,inst2]{Gian Antonio Susto}

\affiliation[inst2]{organization={Human Inspired Technology Research Centre, University of Padova},
            addressline={Via Luigi Luzzatti 4}, 
            city={Padova},
            postcode={35121}, 
            state={PD},
            country={Italy}}

\begin{abstract}

Unsupervised anomaly detection tackles the problem of finding anomalies inside datasets without the labels availability; since data tagging is typically hard or expensive to obtain, such approaches have seen huge applicability in recent years. In this context, Isolation Forest is a popular algorithm able to define an anomaly score by means of an ensemble of peculiar trees called isolation trees. These are built using a random partitioning procedure that is extremely fast and cheap to train. 
However, we find that the standard algorithm might be improved in terms of memory requirements, latency and performances; this is of particular importance in low resources scenarios and in TinyML implementations on ultra-constrained microprocessors. Moreover, Anomaly Detection approaches currently do not take advantage of weak supervisions: being typically consumed in Decision Support Systems, feedback from the users, even if rare, can be a valuable source of information that is currently unexplored. Beside showing iForest training limitations, we propose here TiWS-iForest, an approach that, by leveraging weak supervision is able to reduce Isolation Forest complexity and to enhance detection performances. 
We showed the effectiveness of TiWS-iForest on real word datasets and we share the code in a public repository to enhance reproducibility. 

\end{abstract}




\end{frontmatter}


\section{Introduction}

In recent years, the cost of sensors and microprocessors (MCUs) have significantly decreased; moreover, the new technological scenarios brought by the Internet of Things (IoT) and Industry 4.0, are pushing to the embedding of such sensors and MCUs in an increasing number of systems and devices with \emph{edge computing} capabilities. On one side, the combined availability of sensing and computational capabilities into local devices paves the way for new applications \cite{pishgoo2021hybrid}, like for example the automatic monitoring of the data sensed by edge computing devices \cite{dutta2021tinyml} by means of Machine Learning (ML) approaches; 
on the other hand, such new scenario inspires the research community towards the development of algorithms able to run ML models onto these ultra-constrained devices \cite{banbury2020benchmarking}.
Low resources ML models need to be as light as possible in order to fit the available memory and to be computed on MCUs, moreover they need to efficiently handle multidimensional data that come from a variety of sensors that might be linked to the board.

Concerning the computing paradigm, at the moment we are witnessing a change in the considered architectures: traditional IoT-designed ML models usually heavily rely on cloud computations with resulting latency, bandwidth and privacy concerns that are nowadays representing in many cases an obstacle to the adoption of such solutions \cite{Schneible2017AnomalyDO}. We now instead see the increase of a new paradigm that comes under the name of \emph{TinyML}, where computations are done on the edge tiny devices like MCUs \cite{huvc2021analysis}. This change allows to drastically reduce the latency and the energy consumption caused by the transmission process, and to send to the cloud only the necessary data packet, enhancing the system security. Unfortunately these improvements come at the cost of stricter constraints on the memory and complexity a model can handle to run on these device: the memory capacity goes from some \emph{gigabytes} (cloud GPUs) to \emph{kilobytes} (MCUs), with coherent computational speed scaling \cite{banbury2021micronets}.

In the context of this work, we focus on a particular ML application, Anomaly Detection (AD), that is gaining increasing attention in past recent years. AD algorithms are particularly useful in order to monitor large amount of data \cite{eiras2019large}, and to provide efficient feedback on the data reliability; these models are specifically designed to find unusual patterns inside data that are generated by complex multidimensional processes. This task is usually \emph{unsupervised}, meaning the labels describing if a sample is anomalous, are few or totally absent. In this context the Isolation Forest (IF) is a very appealing algorithm due to its good detecting performances compared to its algorithmic complexity \cite{rubin2020evaluating}. 
However, in the case of ultra-constrained devices, even small improvements in the algorithm can make a difference: the detection performance is only one of the factors that might be considered in the choice of an algorithm to be run on an edge device; other factor might be memory, latency, computational power and energy cost to run on battery power \cite{rubin2020evaluating}.
 

The goal of this paper is to show an algorithm like Isolation Forest that is very cheap to train, could be shrunken in a way that simultaneously reduces its hardware requirements and increases its performances. This can be achieved adding a \emph{weak} supervision in the form of few labels, allowing the forest to be rearranged \emph{without} a proper retraining; indeed the proposed methodology might be used to train a new forest from scratch or to retrofit a previously trained forest with newly obtained labels, similarly to an Online Learning scenario \cite{ren2021tinyol,das2017incorporating}. 

The relaxation of the unsupervised settings is typically reasonable in the context of Decision Support Systems (DSS). In recent years DSSs became pervasive and today are applied to all fields where complex and delicate decision have to be made to assist human operators in the decision-making process like in medicine \cite{sutton2020overview}, \emph{precision} agriculture \cite{kukar2019agrodss}, energy \cite{nybo2014closing}, environment \cite{corradino2019smart} and security \cite{legg2017human,wang2021lightlog}. These systems are equipped with anomaly detection functionalities that allow to automatically monitor the process, giving feedback and alerting the human user to make a possible action; in this context end-users can provide feedback on anomaly detection module suggestions \cite{sejr2021explainable}, making the weakly-supervised scenario that will be considered in this work reasonable.


Moreover, it has to be taken into account that the development of effective AD model is typically a collaborative and \emph{iterative} process between data scientists (AD developers) and end-users (AD users): the firsts choose the algorithm that best fits the given requirements, while the latter ones evaluate if the model ranks the anomalies according to the their expectations. This is necessary since outliers are not uniquely defined but they depend on the user and the context \cite{sejr2021explainable}, therefore requiring different detection strategies \cite{barbariol2022review}; for a more detailed dissertation we refer the interested reader to \cite{sejr2021explainable}.

In this context, we propose here the \approach algorithms that is intended to dialogue with the DSS, following the TinyML paradigm: the DSS system shares with the model the available annotations provided by the end-user, in a weakly supervised fashion \cite{carcillo2021combining,lesouple2021incorporating}, that can be exploited to enhance performance and reduce complexity. Indeed, by assuming that a first model is trained on a fully unlabelled dataset and put in operation on a DSS, during its lifetime it is not unlikely to collect some weak supervision in the form of few labels that can be used to improve the existing detection algorithm. This allows the model to adapt to the user definition of \emph{true} outliers, giving a more domain-specific prediction of outlierliness \cite{das2016incorporating,das2017incorporating}. To the best of our knowledge, \approach is one of the first approaches in the Isolation Forest literature designed to work in a weakly-supervised scenario and to enforce the user-definition of anomaly in an iterative way. Moreover it is the first that, exploiting the available knowledge, reduces the algorithm complexity in view of edge implementations; to enhance the reproducibility we share the code at \url{https://github.com/tombarba/TinyWeaklyIsolationForest}.

This paper is organized as follows: in Section \ref{sec:isolation_forest} the Isolation Forest algorithm will be described focusing on the algorithmic complexity and the ensemble strategy; the datasets employed to test the proposed strategy will be described in the same Section. Then Section \ref{sec:analysis} will go deeper in the analysis of the algorithm describing some issues related to the standard training procedure and showing many examples. Starting from this point, Section \ref{sec:weakly} will describe a weakly-supervised algorithm whose goal is to overcome the previously mentioned obstacles. Here a number of results concerning the application of this new algorithm to real world datasets will be shown, proving the effectiveness of the proposed approach. In the final Section the work will be summarized and future improvements discussed.

\section{Isolation Forest Algorithm and Testing Datasets}\label{sec:isolation_forest}
Isolation Forest (IF) \cite{liu2012isolation} is an ensemble of binary trees named \emph{isolation trees} since their goal is to isolate data points. This algorithm relies on the assumption that anomalies are few and different from normal points, and that recursive space partitioning should isolate anomalous data points (outliers) in an easier way w.r.t. \emph{normal} data points (inliers). This is done by means of an isolation tree that recursively splits the space, choosing randomly with uniform probability the feature and the threshold where to split the space \cite{tokovarov2021probabilistic}; this process is repeated until every point is isolated in a leaf, or the isolation tree reaches a maximum depth. Since it is reasonable to expect that anomalies are isolated faster than inliers, their path length along the tree should be smaller when compared to normal data points. This allows to define an anomaly score $s(\cdot)$ as:
\[ s(x,n) = 2^{-\frac{E[h(x)]}{c(n)}} \]
where $x$ is the input data point, $c(n)$ is a normalising factor representing the average depth of a binary tree with $n$ samples and $E[h(x)]$ is the expected isolation tree depth reached by the point $x$. Then, in many applications, the anomaly score is transformed in a binary label by means of a threshold $\tau$ (usually 0.5): if the anomaly score associated to a point is higher than the threshold, the point is flagged as an anomaly, otherwise is considered normal.

As previously mentioned, IF is an ensemble of different isolation trees that are constructed in a way to guarantee robustness also in the presence of random choices that are present in the isolation procedure: each isolation tree is trained using a bagging strategy, i.e. by using different sub-samples of the same dataset. Experimentally, the IF authors suggested as a guideline to use $t=100$ trees with sub-samples of $\psi = 256$ points for obtaining a stable estimate $E[h(x)]$ using the sampling mean:
\[ \hat{h}(x) =  \frac{1}{t} \sum^{t}_{i} w_i h_i(x) \]
where every isolation tree is implicitly assumed to be equally informative, and therefore weighted by the same constant value $w_i$ = 1. The aforementioned choices for $t$ and $\psi$ are typically adopted by many authors in the literature and they are the default choice in many libraries implementing IF.

Bagging allows IF to achieve linear time-complexity together with small memory requirements \cite{liu2012isolation}, that are very interesting properties when considering \emph{tiny} implementations on MCUs.
Indeed the time complexities in the training stage is $O(t \psi \log \psi)$ while in the testing is only $O(n t \log \psi)$ where $n$ is the number of tested instances \cite{liu2008isolation}; the memory requirements are bounded by the number of nodes of each isolation tree, that is $2 \psi -1$ and the number of trees in the forest $t$ \cite{liu2012isolation}, therefore are $O(t \psi)$.

In the rest of the paper, the original Isolation Forest algorithm with 100 randomly grown trees will be referred as the \emph{standard} or \emph{original} one.

\subsection{Real world datasets}\label{sec:datasets}
As introduced, in this paper we will analyse the original IF algorithm and the proposed TiWS-iForest in terms of efficiency and detection capabilities; we will make this comparison on several datasets that were retrieved in \cite{Rayana:2016} and that are adaptations of the UCI Machine Learning datasets \cite{Dua:2019} for the Anomaly Detection task. Such datasets consists of labelled data coming from different domains: 
\begin{itemize}
\item biomedical (\emph{annthyroid}, \emph{arrhythmia}, \emph{breastw}, \emph{cardio}, \emph{mammography}, \emph{pima}, \emph{thyroid}, \emph{vertebral});
\item environmental  (\emph{cover}, \emph{ionosphere}, \emph{satellite}, \emph{satimage-2});
\item human language (\emph{letter}, \emph{mnist},\emph{optdigits}, \emph{pendigits}, \emph{pendigits}, \emph{speech}, \emph{vowels});
\item others (\emph{musk}, \emph{shuttle}). 
\end{itemize}

\begin{table}[ht]
    \centering
    \begin{tabular}{lrrrr}
\toprule
{} &  n. data &  n. features &  n. anomalies &  contamination \\
\midrule
annthyroid  &    7200 &           6 &          534 &           7.42 \\
arrhythmia  &     452 &         274 &           66 &          14.60 \\
breastw     &     683 &           9 &          239 &          34.99 \\
cardio      &    1831 &          21 &          176 &           9.61 \\
cover       &  286048 &          10 &         2747 &           0.96 \\
ionosphere  &     351 &          33 &          126 &          35.90 \\
letter      &    1600 &          32 &          100 &           6.25 \\
mammography &   11183 &           6 &          260 &           2.32 \\
mnist       &    7603 &         100 &          700 &           9.21 \\
musk        &    3062 &         166 &           97 &           3.17 \\
optdigits   &    5216 &          64 &          150 &           2.88 \\
pendigits   &    6870 &          16 &          156 &           2.27 \\
pima        &     768 &           8 &          268 &          34.90 \\
satellite   &    6435 &          36 &         2036 &          31.64 \\
satimage-2  &    5803 &          36 &           71 &           1.22 \\
shuttle     &   49097 &           9 &         3511 &           7.15 \\
speech      &    3686 &         400 &           61 &           1.65 \\
thyroid     &    3772 &           6 &           93 &           2.47 \\
vertebral   &     240 &           6 &           30 &          12.50 \\
vowels      &    1456 &          12 &           50 &           3.43 \\
\bottomrule
\end{tabular}
    \caption{Summary of the main characteristics of the real word dataset employed in this paper.}
    \label{tab:dataset_description}
\end{table}

In Table \ref{tab:dataset_description}, the datasets are all summarized by providing the number of samples, features, anomalies and the \emph{contamination}, i.e. the percentage of anomalies inside the datasets. The number of samples varies from few hundreds (\emph{vertebral}) to hundreds of thousands (\emph{cover}), while the number of features starts from 6 to 400 in the \emph{speech} dataset. However, the most important characteristic in this context is the contamination: some datasets have less than 1\% of anomalies (\emph{cover}) and reach above the 35\% (\emph{ionosphere}).  All of them have a number of anomalies that exceeds 30 in order to obtain reliable results in the following analysis.
Moreover, all the datasets contain labelled anomalies that allow to measure \emph{a posteriori} the performance of the detection algorithms: the metric employed in this task is the \emph{average precision}; the average precision
\[ \ap = \sum^{n_{\tau}}_{i} p_{i} (r_{i}-r_{i-1}) \] 
summarises the precision $p_i$ and recall scores $r_i$ that can be obtained varying the classification threshold  $\tau$ and is better than the area under the ROC curve when the dataset is highly unbalanced \cite{saito2015precision}.

\section{Analysis of the \emph{standard} Isolation Forest algorithm}\label{sec:analysis}
In the previous Section, Isolation Forest is described as an ensemble i.e. a collection of \emph{weak} learners (isolation trees) that are trained following a completely \emph{random} procedure. This might induce to believe that all of the learners have similar impact and averaging their contribution is the best possible approach, but is it true? Are the isolation trees similarly informative? 

\begin{figure}
\centering
\subfloat[][Good random isolation: the splitting values lie in between normal and anomalous values.]
   {\includegraphics[width=.75\textwidth]{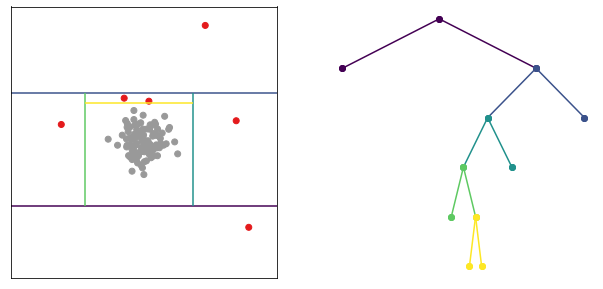}\label{fig:central_cluster_best}} \\
\subfloat[][Bad random isolation: the splitting process focuses on normal data and ignores anomalous ones. This leads to not effective isolation.]
   {\includegraphics[width=.75\textwidth]{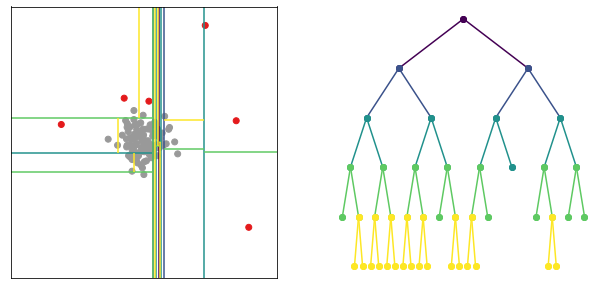}\label{fig:central_cluster_worst}} \\
\caption{Two different isolation trees grown on the same dataset. The anomalies included in the dataset are depicted in red.}
\label{fig:central_cluster}
\end{figure}

A first evidence this is not the case can be seen in Figure \ref{fig:central_cluster_best} and \ref{fig:central_cluster_worst} where two isolation trees are constructed on the same toy dataset composed of a unique central normal cluster and some quite simple anomalies. It is clear that while the first isolation tree makes use of its partitions in an effective way isolating quickly the anomalies, the second focuses too much on normal points and therefore it looses many chances to isolate data.

\begin{figure}
\centering
\subfloat[][Simple dataset made up of two clusters and some scattered anomalies.]
   {\includegraphics[width=.35\textwidth]{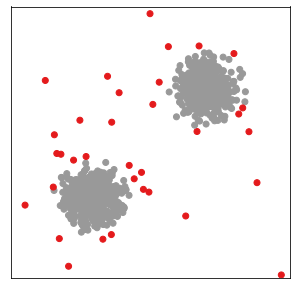}\label{fig:double_cluster}} \qquad
\subfloat[][Histogram of the average precision scores obtained measuring the performances of the isolation trees.]
   {\includegraphics[width=.45\textwidth]{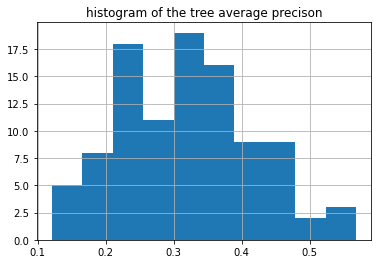}\label{fig:double_cluster_histogram}} 
   \\
\subfloat[][Average precision of different forests built with different strategies.]
   {\includegraphics[width=.55\textwidth]{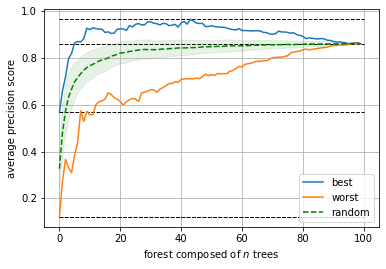}\label{fig:double_cluster_behaviour}} 
\caption{Toy example: double cluster dataset.}
\end{figure}

From the previous qualitative example it seems that not all the isolation trees have the same role in the IF and their equal weighting might not be the best choice.
To get a quantitative feeling of the previous intuition it was settled a more rigorous experiment: 100 isolation trees were randomly grown  following the \emph{standard} procedure, creating the forest named $F_{100}$.  Then each tree was individually tested using the available labels together with the average precision $\ap(\cdot)$ and the histogram of their average precision plotted in Figure \ref{fig:double_cluster_histogram}. From this picture it easy to see that not all the isolation trees behave in the same way: the range between the best and the worst is quite large and the majority lies in between. 

This analysis led us to the following idea, which is at the core of the proposed \approach algorithm: why not to exploit the available weak supervision in order to sort the trees and possibly to get only the best performing isolation trees? This might reduce the model complexity, following the TinyML paradigm, and fine-tune the detection algorithm towards the outlier definition expected by the end-user of the DSS.

With this idea in mind, we made additional analysis by sorting the isolation trees according to three ordering strategies: i) the \emph{best} strategy, the \emph{worst} strategy and the \emph{random} strategy that will be described in the next few lines.
The \emph{best} consists in sorting the isolation trees according to their $\ap$ score in descending order i.e. from the best tree to the worst; the \emph{worst} strategy is the opposite and sorts the trees from the worst to the best. The \emph{random} instead, chooses a random permutation of the trees and therefore it simply shuffles them.

Using the \emph{best} strategy, 100 different isolation forests were built using an increasing number of trees: the first forest contained only the best isolation tree, the second one only the \emph{two} best trees and so on, until the 100-th forest contained all the 100 trees like the \emph{standard} Isolation Forest. At each iteration the average precision of the forest was measured, leading to the blue line in Figure \ref{fig:double_cluster_behaviour}. 
The same was performed with the \emph{worst} strategy (orange line) and with the \emph{random} (green dashed line); however, since the \emph{random} permutation is a non-deterministic strategy, it was repeated 100 times in order to get stable results and to draw the green area.

Figure \ref{fig:double_cluster_behaviour} shows quite interesting results that anticipate some aspects that are also visible in real world datasets: the first and most evident is that good performances can be reached with just 5-40 isolation trees instead of 100. This means that many isolation trees are just overabundant or little informative.
The three lines obviously terminate with the same value (the \emph{standard} Isolation Forest performances) but follow quite different paths: the blue line starts very well and rapidly reaches $\ap_{100}$ while the others require many isolation trees to reach appropriate average precision. 

\begin{figure}
\centering
\subfloat[][Normal data organized in a square toroid and anomalous points inside.]
   {\includegraphics[width=.35\textwidth]{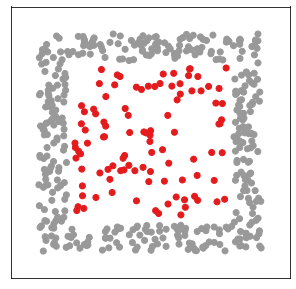}} \quad
\subfloat[][Histogram of the average precision scores obtained measuring the performances of the isolation trees. There are many very bad trees and some few good ones.]
   {\includegraphics[width=.45\textwidth]{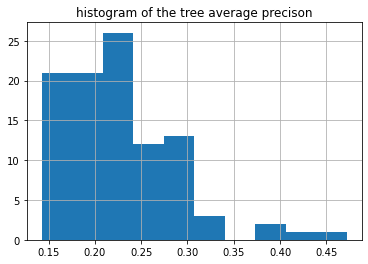}\label{fig:toroid_histogram}} \\
\subfloat[][Average precision of different forests built with different strategies.]
   {\includegraphics[width=.55\textwidth]{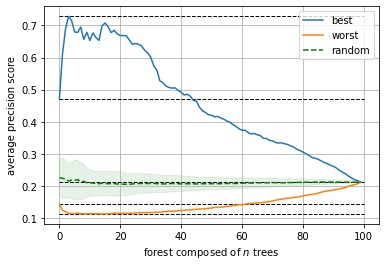}\label{fig:toroid_behaviour}} 
\caption{Toy dataset: square cluster. A more complex example where the advantages of carefully choosing the best isolation trees makes a huge difference in performances.}
\label{fig:toroid}
\end{figure}

Other interesting results can be observed in Figure \ref{fig:toroid} where a second toy dataset is considered: a toroidal dataset with some anomalies in its center; also in this case, the same type of experiments, previously performed on the first toy dataset, were considered, but the previously discussed aspects became even more evident. First of all, the random training procedure generates a lot of very ineffective isolation trees, and very few good ones are present in Figure \ref{fig:toroid_histogram}. This behaviour directly reflects on the IF construction: not only the $standard$ performances were reached very quickly like in the previous toy example, but the best \emph{achievable} performances are much higher than the standard ones. In this example, the best isolation tree alone is better than the whole forest, and with few more trees the forest can reach even better results; indeed on a scale between 0 and 1 with very poor \emph{standard} performances around 0.2, the best achievable performance exceeds 0.7. Unfortunately, due to the corrupting effect introduced by the bad isolation trees on the left of Figure \ref{fig:toroid_histogram}, adding more trees means poisoning the solution leading to quite bad \emph{standard} results (Figure \ref{fig:toroid_behaviour}).
Actually this effect is visible in Figure \ref{fig:double_cluster_behaviour} too, but is less evident and the gap between best performances (blue line) and the mean performances (green line) is smaller. This suggests that in many datasets the IF results may be improved, depending also on the structure of the dataset itself. 

\begin{figure}
\centering
\includegraphics[width=1\textwidth]{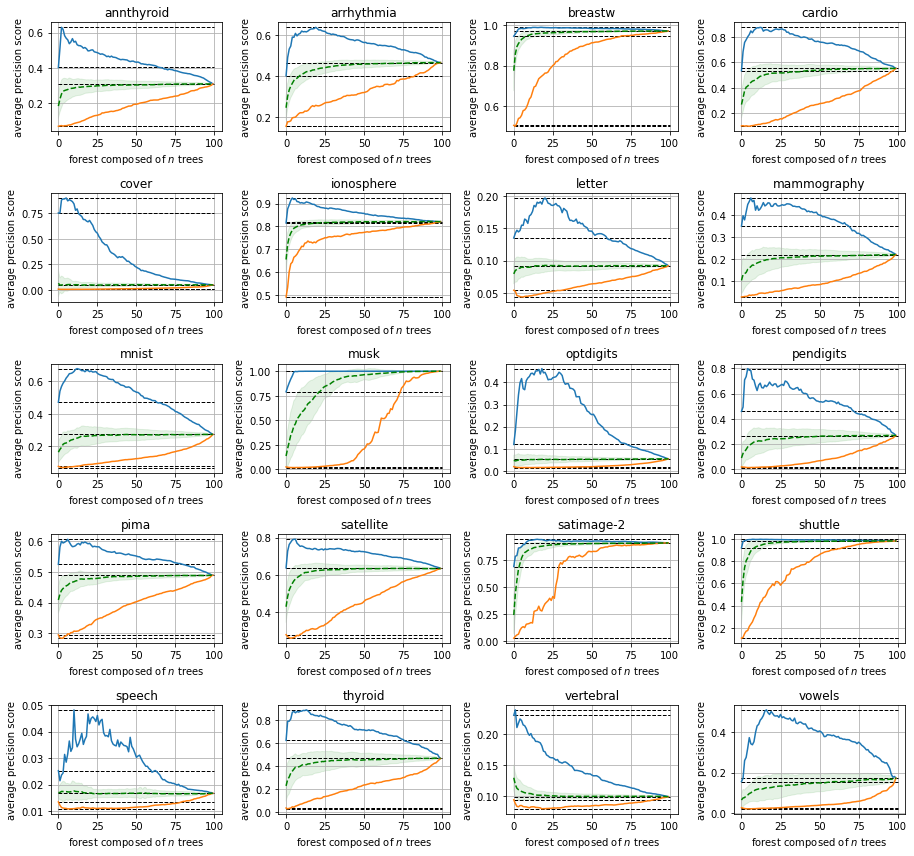}
\caption{Real-world examples. It can be seen how in some cases the gap between the random results (green
distribution) and best achievable (blue line) is quite large, meaning that there is a lot of
room to improve the original algorithm.}
\label{fig:real_world_datasets}
\end{figure}

Repeating the same procedure with some of the most used dataset (the benchmark datasets described in Section \ref{sec:datasets}) it is possible to see similar results (reported in Figure \ref{fig:real_world_datasets}) that can be connected to the examples discussed so far: there seems to be two peculiar behaviours, one described by a logarithmic-shape curve (for example \emph{breastw} and \emph{satimage-2}) and the other by a bell-shape (like \emph{cardio} or \emph{vowels}). Both of them reach the best performance very quickly, i.e. with 5-20 isolation trees, but the bell-shaped starts with higher performances than the standard, reaching very fast the best \emph{achievable} scores and then degrading. In this case, the gap between best and average results (blue and green lines) is very large, suggesting the  \emph{standard} IF algorithm might have a lot of room for improvements. It is not clear the underling motivation of these behaviours: we expect them to be dependent to the structure of the dataset and the definition of outlier that, as explained in the Introduction, it is very dependent on the domain and the end-user expectations.

\section{Weakly-supervised algorithm}\label{sec:weakly}

One may ask how the analysis reported in the previous Section can be exploited in order to improve the original algorithm, and if these results are just over-fitted, meaning the best isolation trees here obtained are valid only for the employed data, or can be generalized to new data points. In other words, can the procedure that selects the best isolation trees lead to over-fitting results? Or, on the contrary, might such procedure be used to learn the best performing trees on a portion of the dataset and to apply those trees on other data coming from the same dataset distribution? This procedure might help in many ways: it can be used to reduce the number of trees and therefore the memory and power consumption of the algorithm, but it might also  be used to increase the average performance of the forest.

\begin{figure}
\centering
\includegraphics[width=1\textwidth]{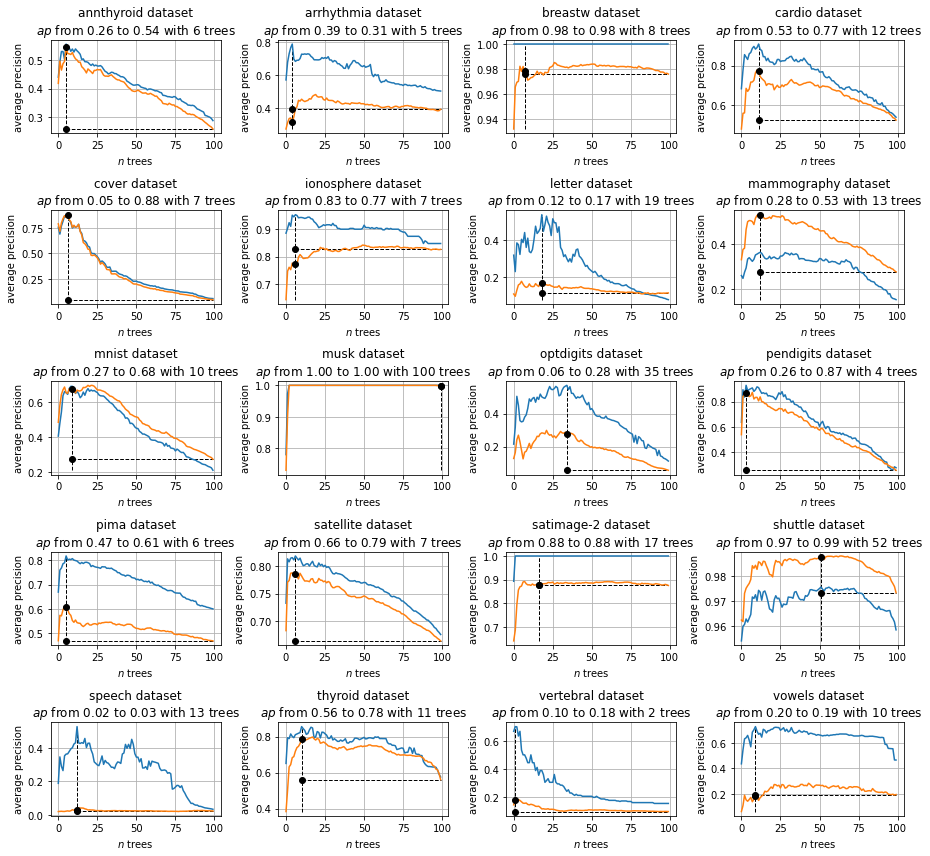}
\caption{Application of the proposed algorithm to the real word datasets, with supervision applied to 20\% of the training dataset. The blue curve shows the performance of the algorithm on the supervised part of the training dataset, while the orange are obtained on the test dataset (50\% of the full dataset). The two black dots highlight: the average precision obtained on the test set obtained with the forest that maximises the average precision during training, the number of trees needed to get these performances, and the average precision of the \emph{standard} algorithm during test.}
\label{fig:real_world_datasets_train_test}
\end{figure}

Based on these questions, a new weakly-supervised algorithm, called \approach, was designed and tested on multiple real word datasets. This is made up of two parts: the \emph{unsupervised} training of the Isolation Forest, and the \emph{supervised} choice of the best performing forest, using the few available labels provided by the domain expert or the DSS. In practice, \approach consists of the following steps:
\begin{itemize}
    \item[i)] training of a standard Isolation Forest on the full training dataset. Here 100 trees are randomly grown without any supervision;
    \item[ii)] using the \emph{supervised portion} of dataset, the previously grown trees are measured according to the average precision metric and sorted accordingly with the descending order;
    \item[iii)] 100 forests are created according to the best isolation tree choice previously discussed i.e. using for each IF the first $i$ best isolation trees. The $i$-th IF will therefore be composed of $i$ isolation trees;
    \item[iv)] the average precision of each of these 100 IF is measured on the same supervised portion of the dataset.
    \item[v)] the forest having the highest average precision is selected. If more than two forests reach the highest precision, the forest with less trees is discarded in favour of the largest; this choice to select the most robust forest among the best ones. 
\end{itemize}

The proposed algorithm was tested over multiple datasets, splitting the full dataset into two equally large sets keeping constant the outlier contamination. Then, to simulate the role of the domain expert or the DSS, a fraction of the training set with the same anomaly contamination was labelled and used in the supervised part of the algorithm.

An example of the results is visible in Figure \ref{fig:real_world_datasets_train_test} where the algorithm is tested on the same datasets shown in Figure \ref{fig:real_world_datasets}, but with a supervised fraction of the training set of about 20\%.
As expected, the reported results resemble the ones shown in Figure \ref{fig:real_world_datasets}, but they are different since now the training algorithm does not see the full dataset (and its anomalies) but a very small portion of those. In this picture the blue line shows the performances obtained from each forest in the supervised portion of the dataset, and the orange line represents the average precision that is measured using the same forests but over the testing set. The vertical line lies in correspondence with the maximal point reached by the blue curve, while the horizontal highlights the performances obtained by the standard forest on the test set. Larger is the interval between the two black dots, better is performing the proposed algorithm. 
The first aspect that needs to be discussed is the performances of the \emph{last} forest, i.e. the last point of the curves in Figure \ref{fig:real_world_datasets_train_test}, that depicts the average performances of the \emph{standard} algorithm in the training (blue) and testing phase (orange); sometimes they perfectly overlap but unfortunately this not always happen due to the presence of different kind of anomalies inside the two datasets, since they are very few and different with respect to the whole dataset. This therefore justifies the different shapes of the training and testing curves that are very often, but not always, similar. The value of the proposed approach lies in the fact that choosing the best trees in the supervised set leads to maximise the chances to get a forest that outperforms the \emph{standard} one even in the test set, and this is clearly visible in Figure \ref{fig:real_world_datasets_train_test}. The only dataset where \approach  fails is the \emph{speech} dataset that has very high dimensionality (400 features) compared with the contamination (1.65\%). On the contrary, the algorithm very often is able to choose a IF that reaches better performances with respect to the standard ones and even in many cases it selects a point that is close to optimal.

\begin{figure}
\centering
\includegraphics[width=0.95\textwidth]{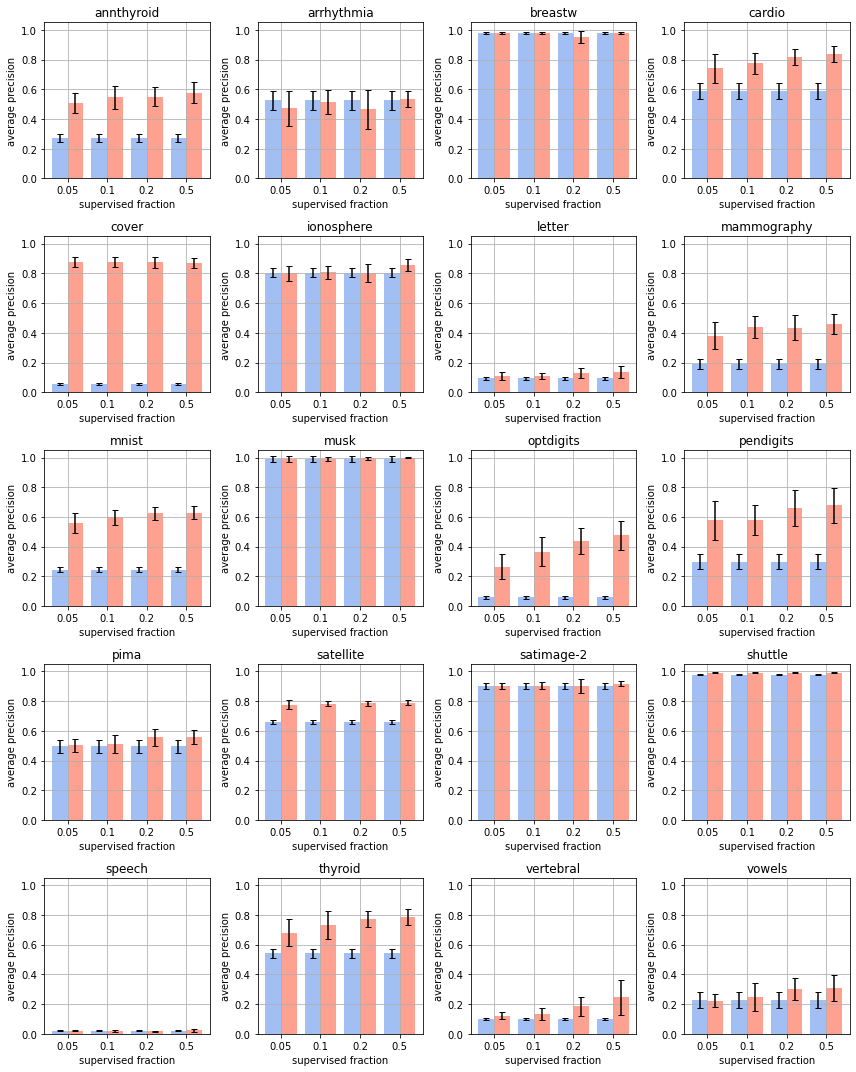}
\caption{Detection performances: results obtained on 24 real word datasets; for each dataset, four different training size percentages are analyzed: the algorithm seems to be robust even with very small training sets.  The average precision of the forest without \approach are depicted in blue while the red represents the improvement due to the described strategy.} \label{fig:results_improvement}
\end{figure}

\begin{figure}
\centering  
\includegraphics[width=0.95\textwidth]{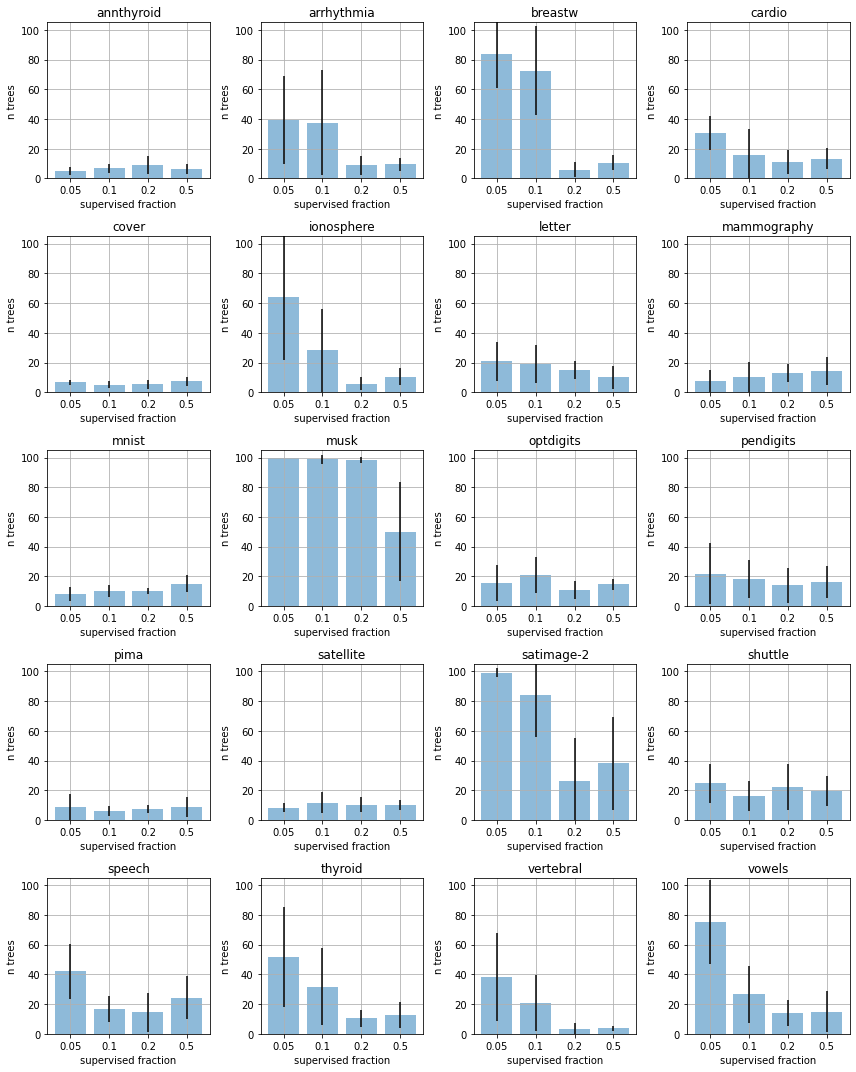}
\caption{Reduction performances: results obtained on 24 real word datasets; for each dataset, four different training size percentages are analyzed: the algorithm seems to be robust even with very small training sets.  The bars represent the number of trees that the proposed strategy needs to reach its performances. As opposed to the standard algorithms, the described one needs a small fraction of memory and computations.}
\label{fig:results_tree}
\end{figure}

Obviously, the previously mentioned Figure \ref{fig:real_world_datasets_train_test} proved the \approach validity, but it needs to be repeated to get robust results and to better quantify the improvements lead by this choice of trees: since anomalies are few and different a particular split of the data may affect the results; to filter out the randomness introduced by this aspects, the algorithm is tested with 10 repetitions, with varying fraction of the labelled training set, and the results are shown in  Figure \ref{fig:results_improvement} and \ref{fig:results_tree}. During the experiments, three values are collected: the baseline i.e. the value of the standard forest on the test set, the number of best trees learnt during training and the average precision measured in the test phase using this subset of trees.

The detection results are depicted in Figure  \ref{fig:results_improvement}, while in Figure \ref{fig:results_tree} the memory savings due to the forest reduction can be noticed by means of the tree cardinality of the selected forest. In Figure \ref{fig:results_improvement} the value of the standard forest on the test set is drawn using the blue color while the red represents the improvement due to the proposed strategy. Even if these can be appreciated in almost every dataset, they are most evident where the standard algorithm behaves in an intermediate way, that is where there is more space for improvement.
The biggest \emph{absolute} improvement is measured on the \emph{cover} dataset, where the average precision goes from about 0.1 to 0.8, probably due to the huge quantity of available data and repeatable anomalies. However, the second biggest \emph{relative} improvement is on the \emph{optdigits} dataset that does not have any special property with respect to the other datasets.

Looking at Figure \ref{fig:results_tree} instead it is possible to see the number of trees that the forest needs to get the results shown in Figure \ref{fig:results_improvement}. The model reduction is most of the times very large, indeed the bars seldom exceed 20-40, saving a lot of memory and computational power since, as  previously explained, both memory requirements and time complexity scales linearly with the number of trees. It is interesting to note that as the fraction of labelled data increases, the algorithm is able to reduce more the size of the selected forest.

The algorithm moreover seems to tolerate quite small training sizes: in all the cases except the \emph{arrhythmia} and the \emph{vowels} dataset, the improvements are present even in experiments with only the 5\% the dataset labelled, and as expected they increase as the train set becomes bigger.

\section{Conclusion}
The detection of anomalies is a critical task in many real life scenarios, however the majority of algorithms are not designed to run on edge devices, learning from unsupervised data and getting the most out of few labelled data, when available. This paper tried to cope with these challenges, primarily observing that even one of the most popular and effective algorithm like the Isolation Forest can be improved considering this scenario. 
This is due to the creation of randomly grown isolation trees that, even tough they are the key aspect of the original algorithm being very cheap to train, some of them risk to damage the solution accidentally. This work tries to solve this problem, designing a simple solution to remove the unnecessary trees, keeping only the most informative with the aid of few labels, named \approach.
As shown in experiments on real world datasets, the forest highly benefits from this procedure and allows the practitioner to include some information in the unsupervised algorithm without a retraining procedure. Not only the detection performances increase, but also the memory and computational cost highly decreases, allowing the implementation of even more constrained devices.  

A similar approach might be used vice-versa to generate the maximum number of isolation trees that fits into a given memory, that are the best with respect to the available supervised data, or it can be applied to other variants of Isolation Forest, like Extended Isolation Forest \cite{hariri2019extended} and Isolation Mondrian Forests \cite{ma2020isolation}. As future works, we will investigate optimized procedure to speed up the search of the TiWSiF solution.

\section*{Acknowledgement}
This work has been supported by MIUR (Italian Minister for Education) under the initiative “Departments of Excellence” (Law 232/2016) and by "Black-box Anomaly Detection: Advanced Approaches and Applications - BADA$^3$" funded by the Department of Information Engineering of University of Padova.

\appendix

 \bibliographystyle{elsarticle-num} 
 \bibliography{bibliography.bib}





\end{document}